\documentclass[letterpaper, 10 pt, journal, twoside]{IEEEtran}  

\usepackage{graphics} 
\usepackage{times} 
\usepackage{amsmath} 
\usepackage{amssymb}  
\usepackage{bm}
\usepackage{todonotes}
\usepackage{multirow}
\usepackage{hhline}
\usepackage{float}
\usepackage{tabularx}
\usepackage{booktabs}
\usepackage{subcaption}
\renewcommand{\vec}[1]{\bm{#1}}
\definecolor{blockcolor}{RGB}{173, 216, 230}
\definecolor{envforcecolor}{RGB}{254, 30, 31}
\definecolor{toolforcecolor}{RGB}{4, 127, 16}
\definecolor{forcepurple}{RGB}{162, 68, 154}
\definecolor{pointgreen}{RGB}{57, 135, 71}

\usepackage{hyperref}

\newcolumntype{C}[1]{>{\centering\arraybackslash}p{#1}}

\begin{document}

\title{Estimating Deformable-Rigid Contact Interactions for a Deformable Tool via Learning and Model-Based Optimization}

\author{Mark Van der Merwe, Miquel Oller, Dmitry Berenson, and Nima Fazeli$^{1}$
\thanks{This work was supported in part by the Office of Naval Research
Grant N00014-24-1-2036 and NSF grants IIS-2113401, IIS-2337870, and IIS-2220876.} 
\thanks{$^{1}$Robotics Department,
        University of Michigan, Ann Arbor, MI, USA 48109
        {\tt\footnotesize \{markvdm, oller, dmitryb, nfz\}@umich.edu}
        }%
}


\maketitle

\begin{abstract}

Dexterous manipulation requires careful reasoning over extrinsic contacts. The prevalence of deforming tools in human environments, the use of deformable sensors, and the increasing number of soft robots yields a need for approaches that enable dexterous manipulation through contact reasoning where not all contacts are well characterized by classical rigid body contact models. Here, we consider the case of a deforming tool dexterously manipulating a rigid object. We propose a hybrid learning and first-principles approach to the modeling of simultaneous motion and force transfer of tools and objects. The learned module is responsible for jointly estimating the rigid object's motion and the deformable tool's imparted contact forces. We then propose a Contact Quadratic Program to recover forces between the environment and object subject to quasi-static equilibrium and Coulomb friction. The results is a system capable of modeling both intrinsic and extrinsic motions, contacts, and forces during dexterous deformable manipulation. We train our method in simulation and show that our method outperforms baselines under varying block geometries and physical properties, during pushing and pivoting manipulations, and demonstrate transfer to real world interactions. Video results can be found at \href{https://deform-rigid-contact.github.io/}{https://deform-rigid-contact.github.io/}.

\end{abstract}



\section{Introduction}
\label{sec:introduction}

\IEEEPARstart{R}{easoning} over the contact between a tool or end-effector and a target object is crucial to enabling performant, safe, and autonomous robotic manipulation. This extends naturally to the case where the contacts are not all well represented by rigid-body contact models. Deformable tools and objects are commonly found in human environments~\cite{yin2021dom} and the inherent compliance may be advantageous for both maintaining contact and preventing excessive forces while interacting with, securing and controlling objects. Additionally, the advent of deformable tactile sensors ~\cite{donlon2018gelslim, alspach2019softbubble, sun2022soft} and soft robot manipulators~\cite{yoo2024moe} introduce new cases of non-rigid contact into robotic manipulation.

In this work, we consider an elastically deformable tool manipulating a rigid object, supported by the environment. This scenario could arise when utilizing deforming tools such as spatula or sponges~\cite{pmlr-v205-merwe23a} or manipulating objects with a soft robot~\cite{punyo} or end-effector~\cite{pmlr-v205-oller23a}. Manipulation of such a system presents two important challenges: First, the deformation of the tool and motion of the rigid object are inherently linked - solving for one requires reasoning about both. However, modeling the deformation of the tool is challenging, due to the high-dimensional and non-linear dynamics~\cite{wi2022virdo}. Second, the system is only partially observable, and we must rely on sensing to intuit deformations and contact forces.

\begin{figure}
    \centering
    \includegraphics[width=\linewidth]{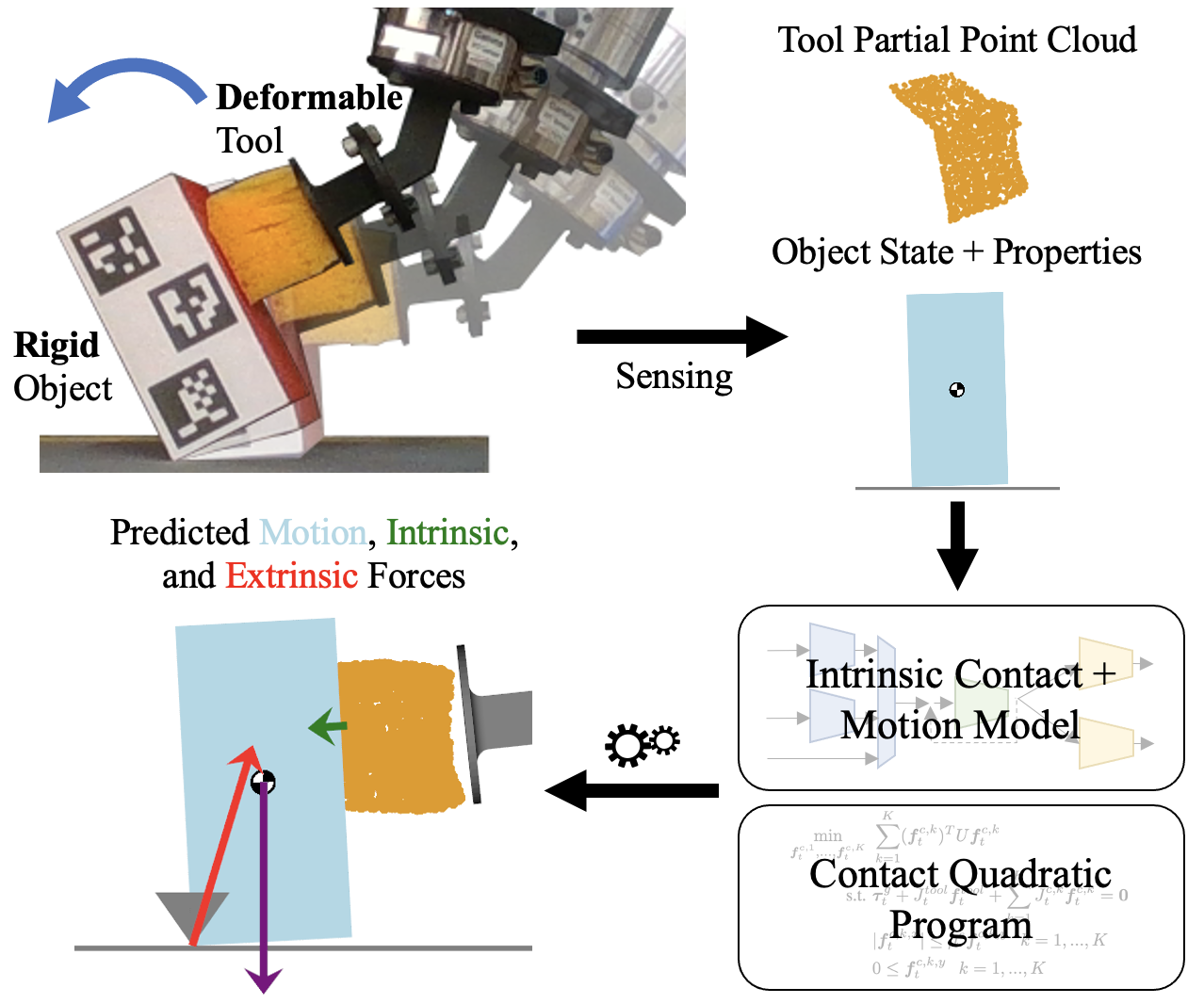}
    \caption{We present a method that estimates motions and forces during dexterous manipulation with a deformable tool. Our proposed method takes in object information (geometry, center of mass, mass, and friction) and sensing from the deforming tool (partial point cloud). It estimates the \textcolor{blockcolor}{block motion}, \textcolor{toolforcecolor}{tool force} the deforming tool enacts on the block, and the \textcolor{envforcecolor}{rigid contact forces} between block and (known) environment, given the robot actions. On the bottom left, we show one-step predictions from our method for a real pivoting execution.}
    \label{fig:intro_figure}
\end{figure}

To address these challenges, we propose a hybrid learning and first principles method for modeling the motion, contacts, and forces on the tool and extrinsic object during manipulation. Accurate motion models are crucial for driving the object toward goal configurations while recovering forces can help enforce desired contact modes and prevent excessive force transfer. 

Our method leverages learning to address the complexity of the paired object motion and deformable tool contacts. It then turns to first-principles and physical priors to recover the extrinsic contacts and forces, enforcing quasi-static motion and Coulomb friction. This allows our method to overcome modeling challenges for the deformable object, while exploiting physical priors for modeling efficiency. We train our method on interactions between a tool and a variety of object geometries and physical properties using simulation~\cite{han2023convex}. We benchmark our method against baselines for modeling block motion and for recovering the environment-object forces. Finally, we deploy our model on a real robot, demonstrating sim-to-real transfer. In summary, our contributions are:
\begin{itemize}
    \item a learning architecture for jointly resolving object motions and tool contacts and forces, given partial sensing and object information.
    \item a first-principles Contact Quadratic Program (CQP) that resolves environment contacts and forces, subject to quasi-static equilibrium and Coloumb friction.
\end{itemize}

\section{Related Work} \label{sec:related_work}

Enabling extrinsic dexterity of freely moving rigid bodies has long been a subject of research, from pushing~\cite{yu2016push} to pivoting~\cite{hou2018pivot} to chaining of multiple manipulating primitives~\cite{hogan2020tactile}. 
Existing work on extrinsic dexterity largely assumes rigid-to-rigid interactions~\cite{hou2018pivot,shirai2023tactile,Oller-RSS-24} and/or assumes sensing at the contact interface~\cite{hogan2020tactile, suresh2024neuralfeels}. We instead investigate deformable-to-rigid contact, where the robot directly controls the deformable tool and indirectly manipulates the rigid body.

Much research forgoes model-based reasoning in favor of learning rigid-body motion from data. To predict the motion of a rigid body, existing work has investigated directly predicting SE3 transforms of objects~\cite{byravan2017se3,byravan2018se3} or predicting point cloud or mesh vertex motion~\cite{pmlr-v205-seita23a,pmlr-v205-allen23a}. Rigid body contact can yield discontinuities that are challenging for learned models, which generally favor smooth approximations. As such, Pfrommer et al.~\cite{pmlr-v155-pfrommer21a} propose to learn rigid body motion by parameterizing inter-body signed distances and contact Jacobians, enabling analytical physical simulation that can reflect rigid contacts. Our work, in-contrast, decomposes the contacts on the extrinsic rigid object to the deformable-to-rigid contact, which we model by learning to predict contact information, and the rigid-to-rigid contact between the object and environment. We note that our system does assume quasi-static motion, unlike in~\cite{pmlr-v155-pfrommer21a}. Calandra et al.~\cite{calandra2015idc} recovers contact forces and incorporates them into dynamics for a robot arm; we recover contact forces and incorporate them into quasi-static equilibrium for a rigid body. Other work seeks to directly predict the motion of heterogenous materials by applying Graph Neural Networks~\cite{li2019particle}. In contrast, we avoid directly modeling the motion of the deforming tool and only focus on the contact forces and rigid body motion.

Using deformation as an indicator for contact and force information has been demonstrated in continuum robots~\cite{ferguson2024unified, wang2024sensing} and volumetric deformables~\cite{wi2022virdo,Merwe-RSS-23}. In this work, we extend the modeling focus to also include the extrinsic object being manipulated.

\section{Problem Formulation}\label{sec:problem}

Our goal is to reason over a) the motion of the extrinsic rigid object, and b) all the forces acting on it. We consider the case where the motion of the robot and rigid object are in the plane of gravity. We assume the following information from each part of the system:

\begin{enumerate}
    \item \textbf{Object and Support Surface:} We assume knowledge of the mass $m^e$, the center of mass $\vec p^{e,CoM}$, the friction between object and rigid supporting surface $\mu^e$, the geometry of the object, provided as a mesh $\mathcal{M}=(\mathcal{V}, \mathcal{E})$, and the geometry of the support surface. Furthermore, we can track the motion of the rigid object to receive the current object pose $\vec{q}^e_t \in SE(2)$. By our assumed knowledge of the environment geometry, we additionally can use the known object pose and geometry to recover the current contact locations $\vec p^{c,1}_t,...,\vec p^{c,K}_t$.
    \item \textbf{Tool:} We assume access to a \textit{segmented} partial pointcloud of the deformable tool $\vec{P}^{tool}$, provided by our sensors. Recent developments in category-agnostic open world segmentation~\cite{Kirillov_2023_ICCV} make recovering segmentation masks possible on the real system.
\end{enumerate}

Given these inputs and assumptions, and the actions to be taken $\vec{a}_{t:t+H}$ for some horizon $H$, where each action is a planar translation and rotation, we wish to recover the future poses of the object $\vec{q}^e_{t:t+H}$ and all forces acting on the object. We split our consideration of the forces into those from the deforming tool and those from the supporting surface. Contact interface representations have take many forms, including points~\cite{manuelli2016cpf}, lines~\cite{kim2022active,ma2021extrinsic,pmlr-v205-merwe23a} and patches~\cite{higuera2023ncf,Merwe-RSS-23}. Here, the most important feature of the contact interaction is not its particular geometry, but rather its \textit{effect}, namely, the resulting force on the block. As such, we consider a simple contact representation: a summary contact point $\vec p^{tool}_t$ and force $\vec f^{tool}_t$, whose resulting wrench on the block is equivalent to the actual and potentially extended contact. As many points and forces can yield an equivalent wrench on the object, we use the centroid of the contact between the tool and object as $\vec p^{tool}_t$, which prevents ambiguity in the representation. Finally, we wish to also recover the set of contact points $\vec p^{c,1}_t,...,\vec p^{c,K}_t$ and forces $\vec f^{c,1}_t,...,\vec f^{c,K}_t$ between the rigid object and the environment. We assume access to a labeled dataset with the specified inputs and outputs listed above, which we utilize to train our method and the baselines.

\section{Method} \label{sec:method}

\begin{figure*}
    \centering
    \includegraphics[width=0.95\linewidth]{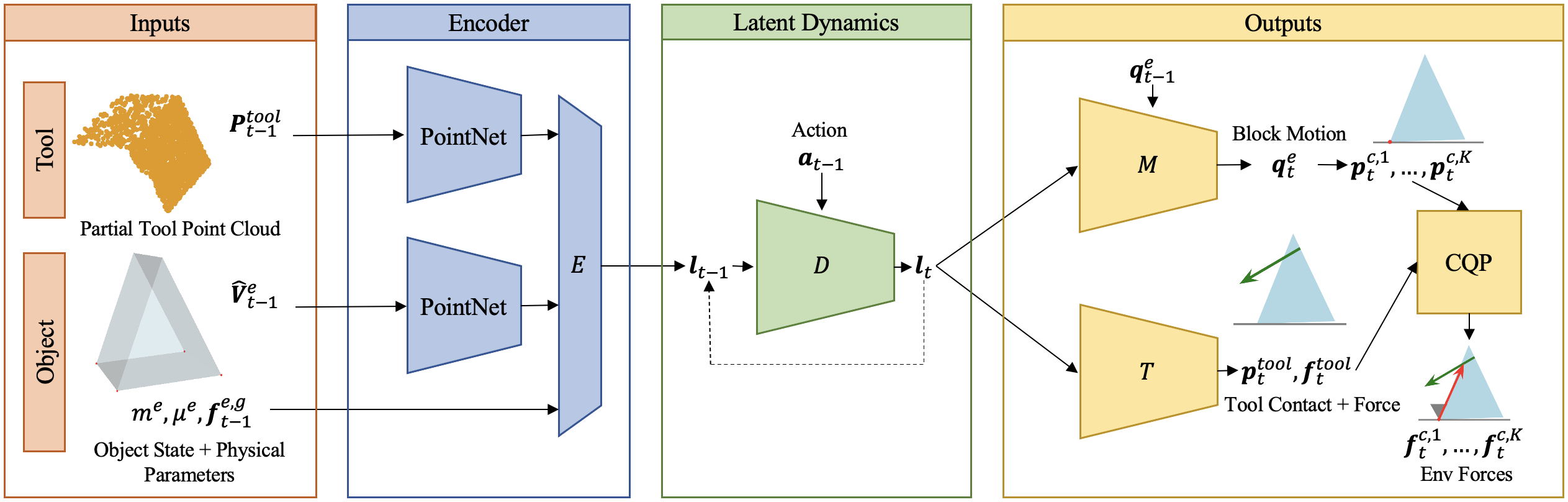}
    \caption{Overview of our proposed method. Our method takes in partial views of the tool via point clouds along with the current object state and parameters and encodes to a learned latent space. Our latent dynamics module then rolls out in the latent space given actions. Finally, our model regresses a) object motion and b) tool summary contact point and force. We use the object motion to determine environment contacts and solve a Contact Quadratic Program (CQP) of our design which resolves the environment contact forces subject to friction and quasi-static equilibrium.}
    \label{fig:method_figure}
\end{figure*}

Our proposed method has two main components - a  learned module for predicting a) the object motion and b) the contact location and forces between the deforming tool and object, and a model-based optimization problem for recovering the resulting frictional contacts with the environment. An overview of our approach can be found in Fig.~\ref{fig:method_figure}.

\subsection{Jointly Learning Object Motion and Deformable-Rigid Contact Interactions}

The underlying mechanics of deforming objects is complex and can be expensive to compute~\cite{yin2021dom,han2023convex}. Simplifying models~\cite{masterjohn2022hydroelastic} can recover forces but lack fidelity to deforming surfaces. Learning-based methods have modeled geometries~\cite{wi2022virdo,pmlr-v205-wi23a} and contact interfaces~\cite{Merwe-RSS-23} and crucially can recover information when only provided with partial information sensing, such as partial pointclouds, but have not recovered forces at the contacting surfaces. We adopt a learning-based approach to be able to recover the complex contacts and motions directly from partial sensing, and directly predict the resulting forces.

\subsubsection{Architecture}

An overview of our proposed architecture is shown in Fig.~\ref{fig:method_figure}. Our architecture design encodes multimodal information from the object and the tool into a learned latent space. A latent dynamics backbone then propagates the latent information provided with the actions. Finally, we have multiple output heads that decode the object motion and tool contact information from the latent. We detail the architecture below.\\
\textbf{Network Inputs:} In order to accurately reason about the interactions, we input multimodal information from the rigid object and the tool. We represent spatial quantities (poses, forces, points, etc.) in the robot end effector frame provided by $\vec{q}^r_t$. We first encode the partial pointcloud of the tool $\vec{P}^{tool}_t$ using a PointNet encoder~\cite{qi2017pointnet}. We encode the geometry and current pose by extracting the mesh vertices $V$ and transforming them to the tracked object location $\vec{q}^e_t$. We additionally use our knowledge of the environment to detect which vertices are contacting the surrounding environment and append these binary contact indicators to each vertex location. We use another PointNet encoder~\cite{qi2017pointnet} to encode the resulting unstructured point set $\hat{\vec{V}}_t^e\in \mathbb{R}^{|V|\times 4}$. Note, that this also allows us to use a single encoder for objects with varying geometries and differing numbers of vertices. We fuse the resulting latent embeddings from the tool and rigid object, along with the object mass $m^e$, object friction $\mu^e$,
and the gravity vector $\vec{f}^{e,g}_t$ expressed in the robot frame. This allows the model to reason about gravitational effects while keeping all learning in the robot frame. These inputs are fused and passed through a final multi-layer perceptron (MLP) to yield the latent code $\vec l_t$. We call this encoder $E$.\\
\textbf{Latent Dynamics:} Given the input actions $\vec a_{t:t+T}$, we rollout the dynamics in the latent space~\cite{hafner2019learning} as the latent space can capture how the resulting deformations, motions, and forces evolve. Our latent dynamics model $D$ is given the current latent state and action and predicts the future latent state.

\begin{equation}
  \vec l_{t+1} = D(\vec l_t, \vec a_t)
\end{equation}

This allows us to recursively predict successive states given action sequences.\\
\textbf{Output Heads:} The final learned components are two output networks which take the latent state at a given time step and estimate the motion. One MLP $M$ is used to estimate the motion of the object, given the current latent state. We predict the motion as a delta motion, utilizing the axis-angle representation~\cite{byravan2017se3,byravan2018se3}.\footnote{Our model can handle full SE(3) poses and actions but all examples considered are planar motions.}
\begin{equation}
  \Delta \vec q^{e}_t = M(\vec l_t)
\end{equation}
These delta poses can be combined with the tracked initial state $\vec q_t^e$ to predict the future rollout states of the rigid object.

A final MLP $T$ is used to directly regress the tool contact point $\vec p_t^{tool}$ and force $\vec f_t^{tool}$.
\begin{equation}
  \vec p_t^{tool}, \vec f_t^{tool} = T(\vec l_t)
\end{equation}

\subsubsection{Training Losses} \label{sec:methodloss}

We assume access to a labeled dataset $\mathcal{D}$ which contains example rollouts of the system with the necessary observations, actions, and labels provided. We train the model performance over a specified horizon $H$. We encode the starting observations using our encoder to yield our starting latent $\vec{l}_t = E(\vec{P}_t^{tool}, \hat{\vec{V}}_t^{e}, m^{e}, \mu^e, \vec{f}^{e,g}_t)$ and recursively apply our dynamics $D$ to yield the sequence of predicted latent vectors $\vec{l}_{t:t+T}$. We then apply our models $M$ and $T$ to recover our estimated object delta poses $\Delta \vec{q}^e_{t:t+H}$, tool contact points $\vec p^{tool}_{t:t+H}$, and tool contact forces $\vec f^{tool}_{t:t+H}$. We then apply the following supervised loss.
\begin{align}
  \mathcal{L}_{pred} =& \sum_{i=t}^{t+H} ||{\Delta \vec{q}_i^{e}}^* - \Delta \vec q_i^e||_2^2 + \alpha ||{\vec p_i^{tool}}^* - \vec p_i^{tool}||_2^2 + \\
  & \ \beta |{|\vec f_i^{tool}}^* - \vec f_i^{tool}||_2^2
\end{align}
Where ground truth values come from our dataset $\mathcal{D}$ and $\alpha$ and $\beta$ are loss weighting terms.

To encourage accurate latent dynamics, we use future observations to encode ground truth latent rollouts. That is we derive $\vec{l}_{t:t+H}^*$ by applying $E$ on the actual observations found when rolling out the system. We then encourage our recursively generated latents to match these values. We additionally regularize the scale of the predicted latent vectors.

\begin{equation}
  \mathcal{L}_{latent} = \sum_{i=t}^{t+H} \gamma || \vec{l}_{t}^* - \vec{l}_t ||_2^2 + \zeta || \vec{l}_t ||_2
\end{equation}

\subsection{Model-Based Estimation of Frictional Object-Environment Contact}

We next turn our attention to estimating the environment contacts. As outlined in Sec.~\ref{sec:problem}, we can recover the set of $K$ active contact points at a given time step $\vec p_t^{c,1},...,\vec p_t^{c,K}$ given our estimated object poses and the known object and environment geometries. The task is to find the contact forces at those points $\vec f_t^{c,1},...,\vec f_t^{c,K}$. To estimate these forces, we exploit our knowledge of Coloumb friction and quasi-static motion to formulate an optimization problem that recovers a set of valid contact forces.

Assuming the system is in quasi-static equilibrium, and given the contact force applied by the tool $\vec f_t^{tool}$, we know that the set of environment contact forces must be the solution to the following Quadratic Program (QP) optimization problem~\cite{Oller-RSS-24}.

\begin{align} \label{eq:cqpstrict}
\begin{split}
  \min_{\vec f_t^{c,1},...,\vec f_t^{c,K}} &\ \sum_{k=1}^{K} (\vec f_t^{c,k})^T U \vec f_t^{c,k} \\
  \text{s.t.} &\ \vec \tau_t^g + J_t^{tool} \vec f_t^{tool} + \sum_{k=1}^K J_t^{c,k}\vec f_t^{c,k} = \vec{0} \\
  &\ |\vec f_t^{c,k,x}| \leq \mu^e \vec f_t^{c,k,y}\ \  k=1,...,K\\
  &\ 0 \leq \vec f_t^{c,k,y}\ \  k=1,...,K
\end{split}
\end{align}

Here $\vec \tau_t^g$ are the gravitational terms, $J_t^{tool}$ and $J_t^{c,k}$ are the contact Jacobians mapping contacts from the contact frame to generalized forces on the object, and we use $x$ and $y$ superscripts to indicate the tangential and normal force components in the contact frame, respectively. The first constraint expresses the quasi-static equilibrium of the rigid object. The second and third constraint enforces Coloumb friction, namely that forces must lie within the friction cone defined by $\mu^e$ and that the forces can only apply pushing forces.

Were the system truly quasi-static and our tool force prediction perfectly accurate, we could utilize Eq.~\ref{eq:cqpstrict} to resolve our environment contact forces. In practice however, error in the model predictions, as well as our training data originating from a system that is not truly quasi-static can make Eq.~\ref{eq:cqpstrict} infeasible. To resolve this without sacrificing accuracy, we propose a relaxation of the QP.

\begin{align} 
\begin{split} \label{eq:cqp}
  \min_{\hat{\vec{f}}_t^{tool},\vec f_t^{c,1},...,\vec f_t^{c,K}} &\ (\Delta \vec f_t^{tool})^T U (\Delta \vec f_t^{tool}) + \rho \sum_{k=1}^{K} (\vec f_t^{c,k})^T U \vec f_t^{c,k} \\
  \text{s.t.} &\ \vec \tau_t^g + J_t^{tool} \hat{\vec{f}}_t^{tool} + \sum_{k=1}^K J_t^{c,k}\vec f_t^{c,k} = \vec{0} \\
  &\ |\vec f_t^{c,k,x}| \leq \mu^e \vec f_t^{c,k,y}\ \ k=1,...,K\\
  &\ 0 \leq \vec f_t^{c,k,y}\ \ k=1,...,K
\end{split}
\end{align}

Here, $\Delta \vec f_t^{tool}=\vec f_t^{tool} - \hat{\vec{f}}_t^{tool}$ In our relaxation, we introduce a new decision variable $\hat{\vec{f}}_t^{tool}$ which we use to achieve force balance. We introduce a cost to minimize the difference between the new decision variable and our original estimated tool force $\vec{f}_t^{tool}$. We interpret the resulting $\hat{\vec{f}}_t^{tool}$ as the closest tool force to the estimated tool force, for which the system is in equilibrium. We utilize the resulting $\vec f_t^{c,1},...,\vec f_t^{c,K}$ as our estimate for the environment forces. We set $\rho$ to a small value (1e-3) to prevent excessive drift from the estimated force while still realizing realistic contact forces.


\section{Implementation} \label{sec:implementation}

\subsection{Data Collection} \label{sec:data}

As described in Sec.~\ref{sec:methodloss}, we train our learned model using supervised learning. This requires us to generate ground truth labels of the contact force. As such, we use simulation to generate example interactions and train our model. We use the Drake simulator~\cite{drake}, as it supports deformable-rigid contacts~\cite{han2023convex}. We attach a simple deformable tool to the end of a gripper. We use a 46mm cube to match our real tool (Fig.~\ref{fig:intro_figure}) and utilize Drake's Finite Element Method simulation, setting Youngs Modulus to 1.1e4, Poisson's ratio to 0.1, and density to $30kg/m^{3}$. For the extrinsic object, we use three object primitives for which we generate variations: a rectangle, triangle and pentagon. All shapes are given a depth of 9cm. For each class, we perform geometric variations as well as varying the mass and friction parameters. The geometry and physical parameter variations used are shown in Table~\ref{tab:object_param_variations}.

\begin{table}
    \centering
    \renewcommand{\arraystretch}{1.0}
    \caption{Object Parameter Variations}
    \begin{tabularx}{\linewidth}{p{1cm}|C{2cm}|C{2cm}|C{2cm}}
    \toprule
    Parameter       & Rectangle & Triangle & Pentagon \\
    \toprule
    \multirow{2}{*}{Geometry} & $w\in [0.05, 0.1]$ & $w\in [0.05, 0.1]$ & \multirow{2}{*}{$l\in [0.05, 0.1]$} \\
    & $h\in [0.08,0.15]$ & $h\in[0.08, 0.15]$ & \\
    \midrule
    Mass & \multicolumn{3}{c}{ $m\in [0.3,0.43]$ kg } \\
    \midrule
    Friction & \multicolumn{3}{c}{ $\mu^e\in \mathcal{B}([0.2, 0.8],\alpha=2,\beta=5)$ } \\
    \bottomrule
    \end{tabularx}
    \label{tab:object_param_variations}
\end{table}


\begin{figure*}
    \centering
    \includegraphics[width=\linewidth]{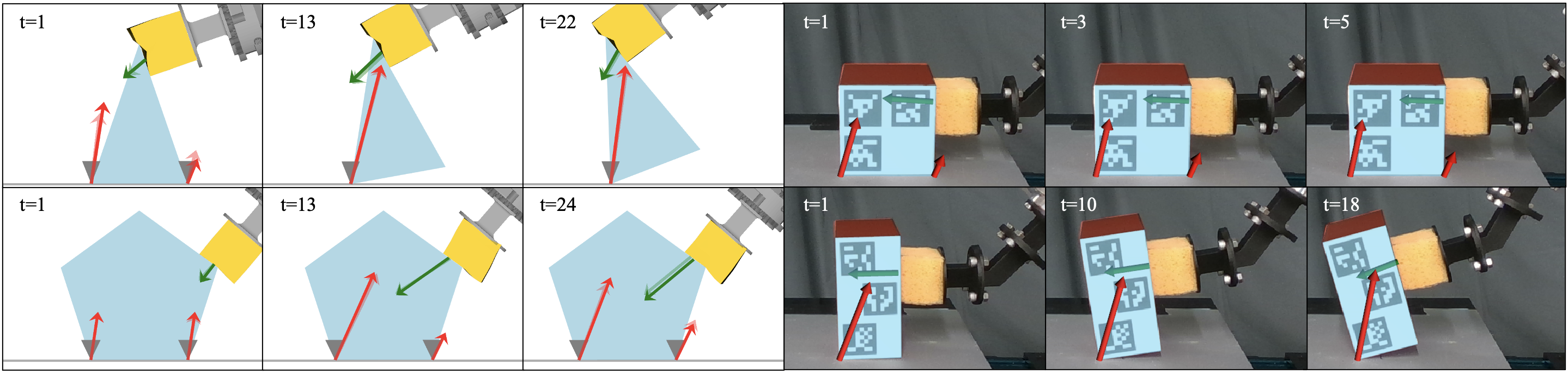}
    \caption{Qualitative Results. \textbf{Left:} We compare our methods one-step predictions (solid) to the ground truth (semi-transparent) \textcolor{blockcolor}{object motion}, \textcolor{toolforcecolor}{tool force}, and \textcolor{envforcecolor}{environment forces} across several manipulation trajectories. The friction cones for each environment contact is shown in gray at the environment point of contact. Our method shows high-fidelity predictions for a) varying object geometries and physical parameters, and b) different interaction primitives (pivoting vs. pushing). \textbf{Right:} We show one-step predictions for real robot executions. The predictions match the observed motion and show qualitatively realistic forces (see Sec.~\ref{sec:real_results} for quantitative results).}
    \label{fig:qual_results}
\end{figure*}

We use a simple scripted policy to collect the data in simulation. We randomize the starting location of the wrist so that initial contact with the object is randomized along the contacting edges. We then perform a simple forward push or a pivot. In the push case, we command a translation along the supporting plane. In the pivot case, we press into the object, then pivot by following an arc around a selected pivot vertex on the rigid object. In both cases, we then add noise to the action to add more variation to the motions. We execute these policies and record the resulting object motion, contact forces from the environment and the tool, and partial point clouds using a simulated camera. We use this to construct our training dataset $\mathcal{D}$.

\subsection{Model Details} \label{sec:model_details}

We collect a training dataset by collecting 5000 push and 5000 pivot trajectories in our simulation with each geometry class we used. Each trajectory samples new geometric and physical properties and runs up to 40 time steps or until the object breaks contact with the tool. We filter the transitions to only include examples in contact. We additionally collect a validation dataset of 500 push and 500 pivot trajectories in the same manner. This yields a final training dataset of 262,780 train transitions and 66,199 validation transitions. During training, we use an action horizon of 4. We implement our network in PyTorch and solve our CQP using CvxPy. Our model is trained for 50 epochs and we use the validation dataset to select the best model. For all experiments, we set our constants from Sec.~\ref{sec:method} as $T=4, \alpha=1.0, \beta=1.0, \gamma=0.1, \zeta=1e-3$.

\section{Results}


\begin{figure}
    \centering
    \includegraphics[width=0.8\linewidth]{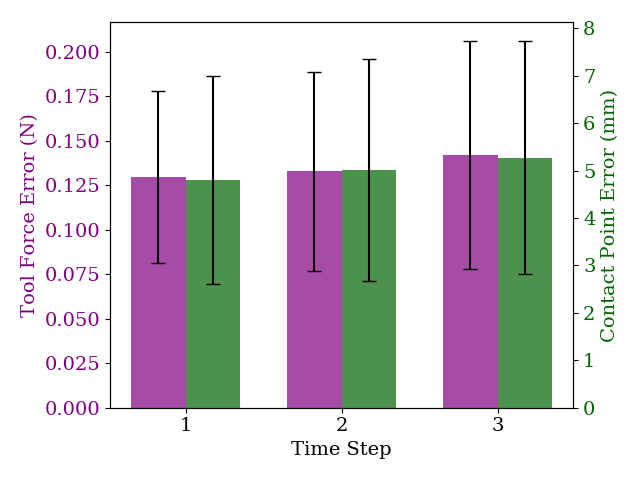}
    \caption{Tool Contact Force (\textcolor{forcepurple}{purple}) and Contact Point Location (\textcolor{pointgreen}{green}) error for our proposed model, plotted by prediction horizon time step. Error bars indicate one half standard deviation.}
    \label{fig:tool_contact_sim}
\end{figure}


\begin{figure*}
    \centering
    \begin{subfigure}[t]{0.32\textwidth}
        \centering
        \includegraphics[width=\textwidth]{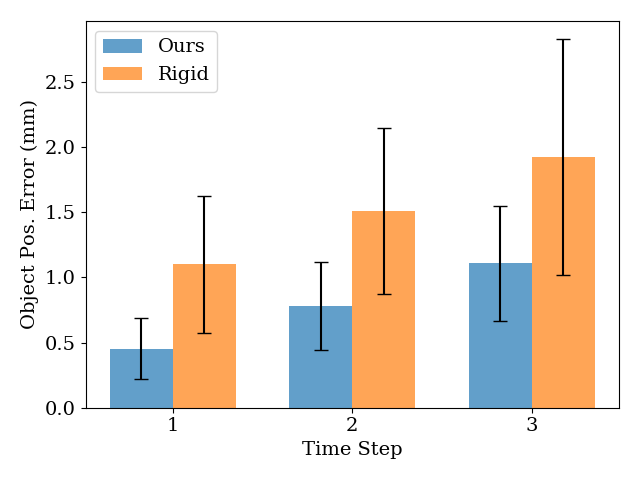}
        \caption{Object Position Error ($\downarrow$)}
        \label{fig:obj_pos_sim}
    \end{subfigure}
    \hfill
    \begin{subfigure}[t]{0.32\textwidth}
        \centering
        \includegraphics[width=\textwidth]{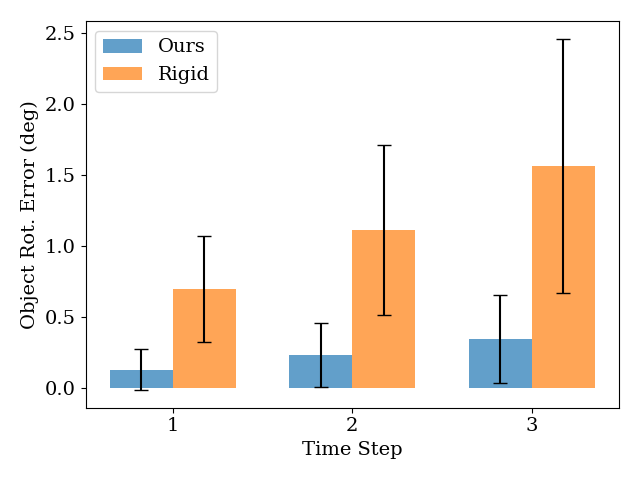}
        \caption{Object Rotation Error ($\downarrow$)}
        \label{fig:obj_rot_sim}
    \end{subfigure}
    \hfill
    \begin{subfigure}[t]{0.32\textwidth}
        \centering
        \includegraphics[width=\textwidth]{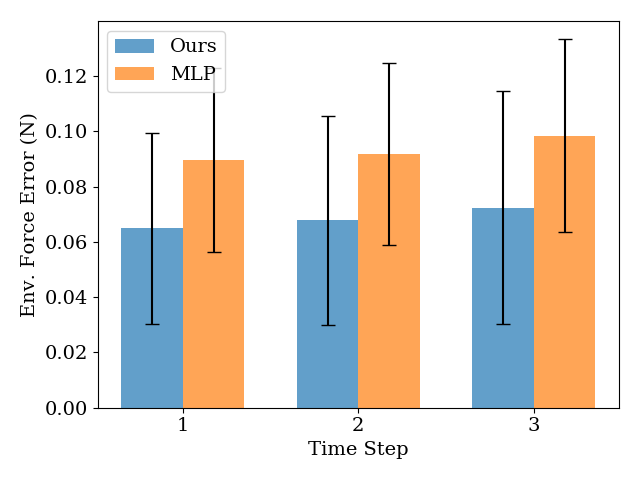}
        \caption{Extrinsic Force Error ($\downarrow$)}
        \label{fig:env_force_sim}
    \end{subfigure}
    \caption{Baseline comparison of our method on test simulated interactions. For (a) and (b) we compare to a Rigid baseline that predicts block motion as if rigidly attached to the tool. For (c) we compare our model-based optimization for extrinsic contact recovery to directly predicting it from an MLP. Error bars indicate one half standard deviation.}
\end{figure*}

\begin{table}
\centering
\renewcommand{\arraystretch}{1.1}
\caption{Real World 1-Step Object and Force Error ($\downarrow$)}
\begin{tabularx}{\linewidth}{p{0.6cm}p{0.3cm}|c|c|c}
\toprule
\multicolumn{2}{X|}{Metric} & Pos. Error (mm) & Rot. Error (deg) & Force Error (N) \\
\toprule
\multirow{2}{*}{Blk. 1} & Pivot & 0.364 (0.128) & 0.262 (0.125) & 0.745 (0.292) \\ 
 \cline{2-5} 
& Push & 1.163 (0.420) & 1.144 (0.613) & 0.977 (0.445) \\ 
 \hline 
\multirow{2}{*}{Blk. 2} & Pivot & 0.466 (0.196) & 0.071 (0.054) & 1.033 (0.382) \\ 
 \cline{2-5} 
& Push & 2.044 (0.211) & 0.069 (0.032) & 1.180 (0.160) \\ 
 \bottomrule
\end{tabularx}
\label{tab:real_env_obj}
\end{table}


\begin{figure}
    \centering
    \includegraphics[width=0.95\linewidth]{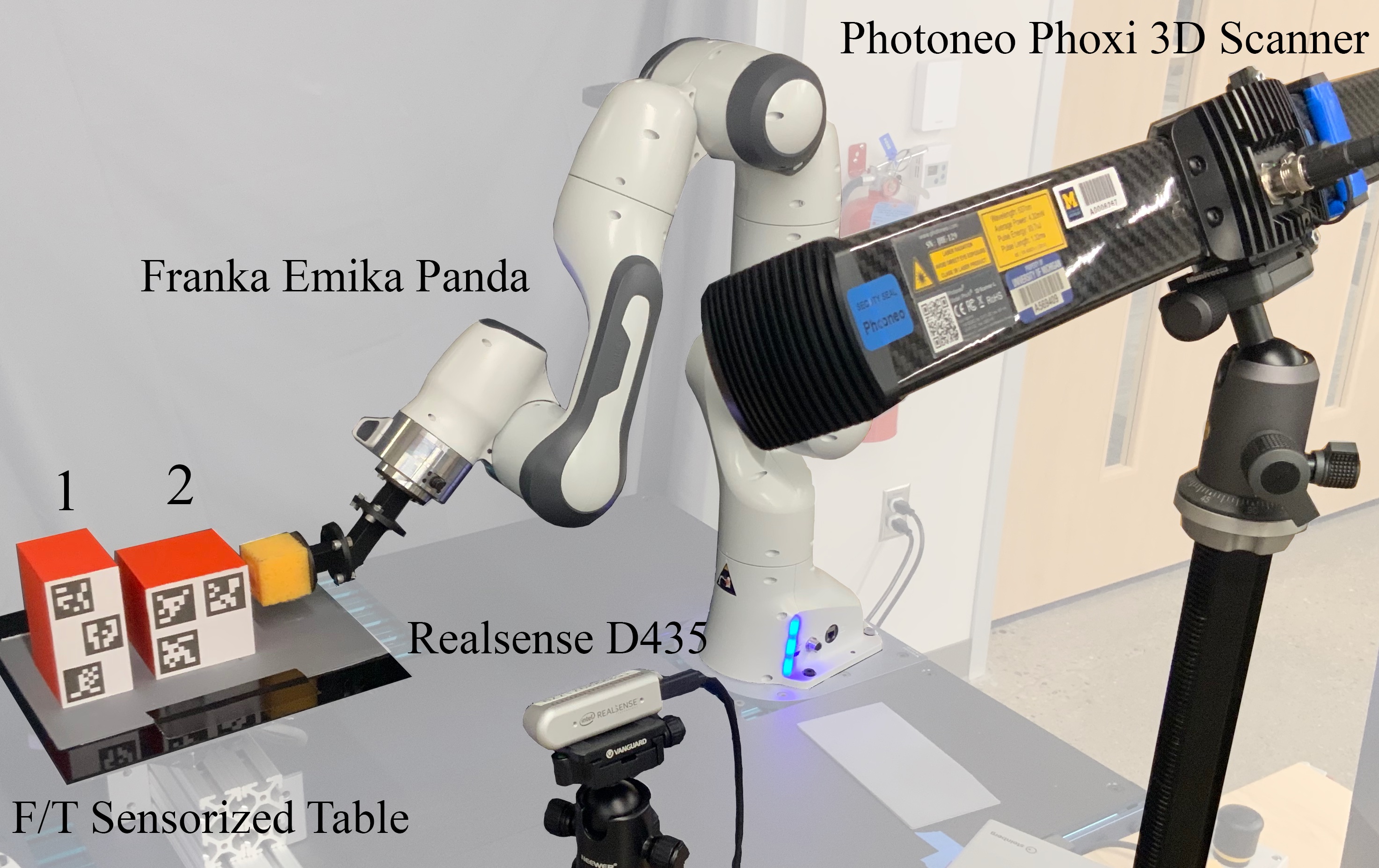}
    \caption{Our real robot setup, with the test objects. We use the Realsense D435 to track the block pose and the Phoxi 3D Scanner to recover partial pointclouds of the tool. We use a Force/Torque sensorized table to indirectly evaluate our force predictions.}
    \label{fig:real-setup}
\end{figure}

\subsection{Simulated Results}

We start by examining our method's performance on a test dataset. We follow the same procedure outlined in Sec.~\ref{sec:implementation} and collect 500 push and 500 pivot trajectories per object class. We examine our methods ability to predict tool contact information and object motion and forces.

\subsubsection{Tool Contact Prediction}

We report our method's mean performance over a prediction horizon of three steps for tool contact point and contact force accuracy across all our simulated test trajectories in Fig.~\ref{fig:tool_contact_sim}. For all interactions and across our prediction horizon, our method showed the ability to accurately estimate future contact points between the deforming tool and extrinsic block, to within 6mm on average (against a tool of size 46mm), and showed the ability to estimate future contact forces to within 0.2N on average.

\subsubsection{Object Motion Prediction}
To benchmark our method's object motion performance, we introduce a \textit{Rigid} baseline which assumes that the block moves ``rigidly'' with respect to the tool. This assumes that the deformation of the tool is negligible for the object motion and doesn't model collisions.

We report our method and the rigid method's object position and rotation accuracy in Figs.~\ref{fig:obj_pos_sim} and~\ref{fig:obj_rot_sim}. We show that across all interaction types, our method outperformed the \textit{Rigid} baseline, with sub-millimeter position accuracy and less than 0.5 degrees of rotational error on average across the full prediction horizon.

\subsubsection{Environment-Object Force Prediction}

A key part of our proposed method is utilizing our physical priors and the quasi-static assumption to solve for the force between the environment and extrinsic object. We hypothesized that this allows for accurate modeling without requiring additional training effort. To benchmark this approach, we create a variation which replaces the CQP with another MLP following the form of our other network output heads. In particular, it takes in the latent state $\vec l_t$ and directly predicts the contact forces $\vec f_t^{c,1},...,\vec f_t^{c,K}$. In practice, the baseline predicts forces at set candidate points on the object and we use the predicted object motion to detect which points lie in contact.

We report our method and baseline method's environment-object force accuracy across all interaction types in Fig.~\ref{fig:env_force_sim}. We find that our proposed method outperforms the baseline, yielding the best average force prediction across all scenarios. Our method consistently performed to within 0.08N of error on average, across the prediction horizon. Our results indicate that our proposed CQP method for recovering environment forces outperforms direct learning. 

Finally, Fig.~\ref{fig:qual_results} (Left) shows qualitative examples of our methods predictions on simulated data with different objects and actions (pivoting and pushing). The predicted object motion, tool contact and force, and environment forces are consistent with the ground truth, as indicated by the significant overlap with the ground truth (shown as semi-transparent).

\subsection{Real Robot Results} \label{sec:real_results}

We demonstrate our methods performance on a real robot system. We use a Franka Emika Panda robot and interact with two rectangular objects. Our setup and the test objects are shown in Fig.~\ref{fig:real-setup}. We collect 5 push and 5 pivot trajectories with each object. We use a Realsense D435 and AprilTags to track the object state and a Photoneo Phoxi 3D Scanner sensor and the Segment Anything Model~\cite{Kirillov_2023_ICCV} to get our segmented partial pointcloud of the tool. Finally, we mount an ATI Gamma Force/Torque sensor under the tabletop. This allows us to compare the cumulative force experienced by the table via the extrinsic contacts, which we compare to our predicted extrinsic contacts.

In Table~\ref{tab:real_env_obj}, we show our one-step object motion prediction and extrinsic force prediction on the real data. Despite sensor noise and the sim-to-real gap, we are still able to predict object motion to within roughly 2mm and 2 degrees error. Our extrinsic contact prediction is within roughly 1N on average. In Fig.~\ref{fig:qual_results} (Right), we show full qualitative predictions of block motion, and intrinsic/extrinsic forces. We see that our method yields qualitatively realistic force estimates given real sensor feedback on the physical system.

\subsection{Force Tracking}

\begin{figure}
    \centering
    \includegraphics[width=\linewidth]{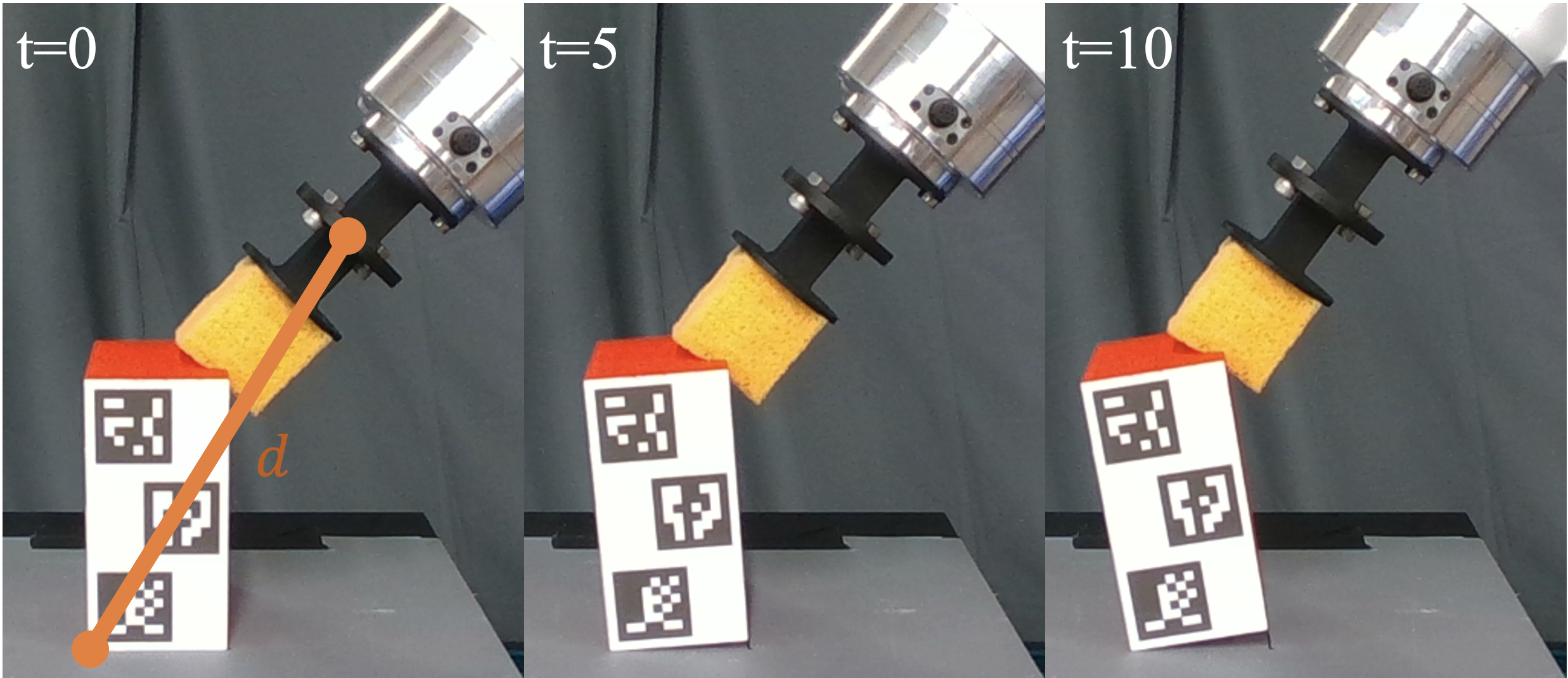}
    \caption{Our pivot force tracking task, with the decision variable $d$ shown in leftmost frame.}
    \label{fig:pivot_plan_vis}
\end{figure}

Finally, we demonstrate a task utilizing our force predictions. Regulating force on the extrinsic object can be important in cases where the object or environment are fragile, such as when handling food, or when the robot or end effector can break or tear. Here, we demonstrate force tracking using our proposed extrinsic contact estimates. We use Block 1 from Sec.~\ref{sec:real_results} and execute a pivoting motion, pivoting around the contact point furthest from the robot. After an initial press into the object, we have the robot follow an arc, such that the robot wrist is a specified distance $d$ from the pivot point. We setup a simple greedy controller which samples various distances $d$ for the next step, thereby allowing it to press harder or release as it pivots. We use our proposed model and predict the motion and forces for the sampled actions (with a single-action horizon). We then select the action with the resulting cumulative force magnitude closest to the desired force setting $\nu$. We set the target force magnitude $\nu$ to be the gravitation force given the mass of the block plus 1.5N. We can then use the F/T sensor mounted on the environment to determine approximately how well we tracked the desired force. An example execution is shown in Fig.~\ref{fig:pivot_plan_vis}.

\begin{figure}
    \centering
    \includegraphics[width=0.9\linewidth]{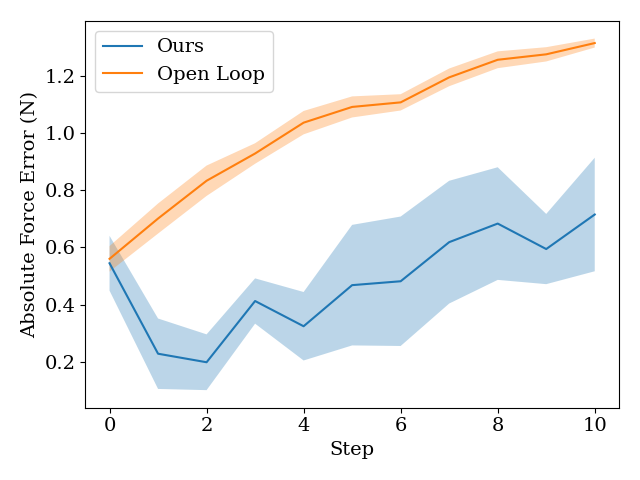}
    \caption{We plot, over five trials, the evolution of the absolute force magnitude error as a function of the trajectory step. Our method is able to better track the desired force, compared to a method without force feedback executing a similar pivot.}
    \label{fig:pivot_plan_force}
\end{figure}

We ran five trials using this simple greedy controller to track the desired force profile. We then executed our baseline, which started at the same starting configuration, but kept the initial $d$ fixed. We used our environment Force/Torque sensor to measure the actual composite force transfer and plot how the errors evolved over the trials in Fig.~\ref{fig:pivot_plan_force}. Our method is able to better track the desired force magnitude, while the open loop plan gradually drifts further from the desired force profile.

\section{Discussion}

We present a method capable of jointly recovering object motion and intrinsic/extrinsic contact information during dexterous deformable manipulations. We found that our first-principles approach outperformed learning methods when recovering environment forces. In future work, we hope to leverage this work for the planning of dexterous deformable motion, where we aim to utilize force estimates to reason about contact modes and force targets.

A limitation of our method is the assumption of quasi-static motion. While this is a common assumption made during dexterous manipulation~\cite{hou2018pivot,Oller-RSS-24}, we may wish for more dynamic motions during certain tasks~\cite{wang_swing_bot_2020}. A potential direction to investigate is incorporating predicted forces into the underlying rigid body dynamics~\cite{calandra2015idc}.

We have limited our investigation here to planar motions of the extrinsic object. While our method can be extended to full SE(3) motions, this requires extending the data collection in simulation to consider those motions and extending our CQP to consider 3D forces.

Additionally, our method assumes knowledge of the geometry of the extrinsic object and the environment, which may not be readily available. One potential remedy would be utilizing a shape reconstruction algorithm to estimate these geometries~\cite{Park_2019_CVPR}.

Finally, our method relies on supervised learning, requiring simulated data to train and successful transfer from simulation to the real robot. Using object motion consistency with predicted forces could be a way to directly train on real system rollouts~\cite{pmlr-v155-pfrommer21a}.




\bibliographystyle{IEEEtran}
\bibliography{main}

\end{document}